\documentclass[letterpaper, 10 pt, conference]{ieeeconf}
\IEEEoverridecommandlockouts                              % This command is only needed if 
\usepackage{graphics} % for pdf, bitmapped graphics files
\usepackage{graphicx}
\usepackage{epsfig} % for postscript graphics files
\usepackage{mathptmx} % assumes new font selection scheme installed
\usepackage{times} % assumes new font selection scheme installed
\usepackage{amsmath} % assumes amsmath package installed
\usepackage{amssymb}  % assumes amsmath package installed
\usepackage[T1]{fontenc}
\usepackage{lmodern}
\usepackage{bm}
\usepackage{bbm}
\newcommand{\assumptions}[1]{\noindent\textbf{#1}}
\usepackage{newtxtext,newtxmath}
\usepackage{mathrsfs}
\usepackage{algorithm, algorithmicx, algpseudocode}
\usepackage[dvipsnames]{xcolor}
\usepackage{dsfont}

\title{\LARGE \bf Set-Membership Estimation for Fault Diagnosis of Nonlinear Systems}

\author{Anastasios Tsolakis$^{1}$, Laura Ferranti$^{1}$, and Vasso Reppa$^{1}$% <-this % stops a space
\thanks{*The research leading to these results has received funding from the European Union’s Horizon 2020 research and innovation program under grant agreement No. 101096923 (SEAMLESS Project). This publication reflects only the authors’ view, exempting the European Union and the granting authority from any liability.}% <-this % stops a space
\thanks{$^{1}$ The authors are with the faculty of Mechanical Engineering, Delft University of Technology, 2628 CD Delft, The Netherlands (e-mail: {\tt\small a.tsolakis@tudelft.nl}).}%
}

\begin{document}

\maketitle
\thispagestyle{empty}
\pagestyle{empty}

%%%%%%%%%%%%%%%%%%%%%%%%%%%%%%%%%%%%%%%%%%%%%%%%%%%%%%%%%%%%%%%%%%%%%%%%%%%%%%%%

\begin{abstract}
This paper introduces a Fault Diagnosis (Detection, Isolation, and Estimation) method using Set-Membership Estimation (SME) designed for a class of nonlinear systems that are linear to the fault parameters. The methodology advances fault diagnosis by continuously evaluating an estimate of the fault parameter and a feasible parameter set where the true fault parameter belongs. Unlike previous SME approaches, in this work, we address nonlinear systems subjected to both input and output uncertainties by utilizing inclusion functions and interval arithmetic. Additionally, we present an approach to outer-approximate the polytopic description of the feasible parameter set by effectively balancing approximation accuracy with computational efficiency resulting in improved fault detectability. Lastly, we introduce adaptive regularization of the parameter estimates to enhance the estimation process when the input-output data are sparse or non-informative, enhancing fault identifiability. We demonstrate the effectiveness of this method in simulations involving an Autonomous Surface Vehicle in both a path-following and a realistic collision avoidance scenario, underscoring its potential to enhance safety and reliability in critical applications. 

\end{abstract}

\section{Introduction}
In recent years, autonomous systems have advanced rapidly, transforming a wide range of human activities such as manufacturing, transportation, agriculture, and environmental monitoring. These technologies promise enhanced efficiency, reduced human error, and the ability to operate in hazardous environments. However, as these systems expand into more sectors, they increasingly depend on sophisticated technology and complex hardware, escalating the intricacies of their operational framework.

Relying on critical components like sensors, actuators, and computational units introduces significant safety and reliability challenges. Faults in these components can lead to failures, posing catastrophic risks and jeopardizing safety. Ensuring safety and reliability in these technologies is of paramount importance, necessitating robust mechanisms to manage and mitigate faults effectively. This paper introduces a method to detect, isolate, and estimate faults, in a wide range of mobile robotic platforms (ground, marine, and aerial vehicles) described as nonlinear mechanical systems that can be expressed linearly to the parameters of interest.

% \subsection{Related Work}
Faults in robotic systems have long been a critical concern across various domains, including ground \cite{karras2020}, marine \cite{cristofaro2014, tsolakis2024b}, aerial \cite{izadi2011, nan2022}, and multi-agent systems \cite{guo2012}. A range of diagnostic and reconfiguration techniques have been deployed, such as observers \cite{cristofaro2014}, Kalman filters \cite{izadi2011}, and learning-based methods \cite{zhang2021}. Faults primarily undermine system performance in two ways: firstly, they can impair the controllability or observability of the system, regardless of the control strategy employed. This typically necessitates redundancy in actuators and sensors to maintain operational capability after a fault has occurred. Secondly, faults create a discrepancy between the system's theoretical model and the actual system itself. If significant, this mismatch can severely compromise controller performance, potentially leading to unpredictable and hazardous behavior.

Fault diagnosis (FD) typically involves the detection, isolation, and estimation of faults, a process that presents significant challenges. The main challenge arises from various sources of uncertainty such as model inaccuracies, environmental disturbances, and measurement noise, which complicate the accurate identification of faults within the system. Set Membership Estimation (SME) has been widely utilized for FD, offering notable advantages. SME eliminates the need for knowing statistical distributions by relying solely on boundedness assumptions. FD through SME facilitates a direct approach by concentrating on fault parameters rather than indirectly inferred residuals, employing inverse tests for fault detection, and concurrently estimating the feasible parameter set from historical input-output data. Specific implementations of SME use zonotopic parameter sets for fault detection, as demonstrated in \cite{puig2010, blesa2011}, and employing ellipsoids to delineate the feasible parameter set, as seen in \cite{reppa2011}, \cite{reppa2016}, \cite{valiauga2021}. Other studies, such as \cite{chabane2016, zhang2024}, apply SME to the system's state for FD. However, the aforementioned works, typically focus on either nonlinear systems without both state and output uncertainties or simpler linear and single-output systems.

Recent work has renewed interest in SME, particularly in adaptive control. Contributions such as \cite{tanaskovic2014} have combined SME with Model Predictive Control (MPC) in a Robust Adaptive MPC (RAMPC) framework, enabling planning based on nominal parameters while maintaining robustness against all feasible parameter realizations. Extensions of SME in RAMPC frameworks for linear systems have been explored in \cite{lorenzen2019, didier2021}, and for nonlinear systems in \cite{kohler2021}, though the challenge of handling both state and output uncertainties simultaneously remains largely unaddressed.

Inspired by SME’s suitability for FD and its compatibility with MPC in a RAMPC framework, this work proposes an FD method based on SME, aimed at enhancing the trajectory optimization method introduced in \cite{tsolakis2024a} to improve safety in environments shared with human-operated vehicles. Specifically, this work extends SME to \textit{nonlinear systems} affected by both \textit{state disturbances} and \textit{measurement noise}—a gap in the current state of the art. The method employs an inverse test for fault detection and isolation, with fault estimation achieved through continuous updates to the feasible parameter set and a fault parameter estimate. The key contributions of this work are:
\begin{itemize}    
    \item Set-membership estimation to nonlinear systems, accounting for both disturbances and measurement noise. This capability ensures \textit{false alarm immunity} by design, thereby increasing the robustness of the fault detection process.
    \item A tighter outer approximation of the feasible parameter set that balances accuracy and computational efficiency, based on user-defined preferences. This leads to improved \textit{fault detectability}, reducing the risk of missed detections and enhancing the system’s responsiveness to faults.
    \item Adaptive regularization in fault parameter estimation to handle cases of sparse, non-informative measurement data, resulting in improved \textit{fault identifiability}.
\end{itemize}

This paper is organized as follows: Section \ref{s:problem_formulation} describes the problem formulation, laying the groundwork for the methodology discussed. Section \ref{s:method} elaborates on our FD method with a detailed analysis of how the different components of SME are derived and used in the FD logic. Section \ref{s:results} demonstrates the application of our methodology through scenarios involving an Autonomous Surface Vehicle (ASV), showcasing the effectiveness of our approach. Section \ref{s:conclusions} concludes with some remarks on the implications of our findings and suggestions for future work.

\section{Problem Formulation}\label{s:problem_formulation}
Consider the following discrete, nonlinear system, which is linear in the vector of fault parameters denoted as $\bm{\theta} \in \mathbb{R}^p$:
\begin{equation}\label{eq:nonlinear_system}
    \bm{z}_{k+1} = \bm{f}(\bm{z}_{k}) + \bm{G}(\bm{u}_{k}) \bm{\theta} + \bm{d}_{k}
\end{equation}
where $\bm{z}_{k} \in \mathbb{R}^n$ is the state, $\bm{u}_{k} \in \mathbb{R}^m$ is the input, and $\bm{d}_{k} \in \mathbb{R}^n$ is the unknown disturbance acting on the system. We assume that the autonomous map $\bm{f}(\cdot) \in \mathbb{R}^n$ and the input map $\bm{G}(\cdot) \in \mathbb{R}^{n \times p}$ are both known. Additionally, we assume that the full state of the system can be measured, albeit corrupted by measurement noise, as:
\begin{equation}\label{eq:state_measurement}
    \bm{y}_{k} = \bm{z}_{k} + \bm{n}_{k}
\end{equation}
where $\bm{y}_{k} \in \mathbb{R}^n$ is the state measurement, and $\bm{n}_{k} \in \mathbb{R}^n$ is the unknown additive measurement noise.

\assumptions{Assumption 1}
The disturbance $\bm{d}_{k}$ and noise $\bm{n}_{k}$ are unknown but bounded signals with known bounds denoted as $\bm{\bar{d}}$ and $\bm{\bar{n}}$, respectively:
\begin{equation}\label{eq:disturbance_bound}
    |\bm{d}_{k}| \leq \bm{\bar{d}}, \quad \forall k = 1, 2, \dots
\end{equation}
\begin{equation}\label{eq:noise_bound}
    |\bm{n}_{k}| \leq \bm{\bar{n}}, \quad \forall k = 1, 2, \dots
\end{equation}
The inequalities between vectors are to be interpreted element-wise, where $|\cdot|$ denotes the matrix modulus function, i.e., the element-wise absolute value.

\assumptions{Assumption 2}  
The fault parameter vector $\bm{\theta} \in \mathbb{R}^p$ is time-invariant, with elements $\theta_i \in [0,1]$, $i = 1, 2, \dots, p$, describing the health of the system. The values of $\theta_i$ represent the following conditions:
\begin{equation}\label{eq:fault}
    \theta_i = 
    \begin{cases}
        \theta_i = 1, \ \forall i \in \{1, 2, \dots, p\}, & \text{healthy system} \\
        \theta_i < 1, \ \exists i \in \{1, 2, \dots, p\}, & \text{faulty system}
    \end{cases}
\end{equation}

The objective of this work is to detect, isolate, and estimate faults with guarantees, using SME to compute an outer approximation of the feasible parameter set and an estimate of the fault parameter. Inverse tests are then applied to detect faults, isolate them, and refine fault parameter estimates upon detection.

\section{Proposed SME Method}\label{s:method}
\begin{figure}[tbh!]
    \centering
    \includegraphics[width=\columnwidth]{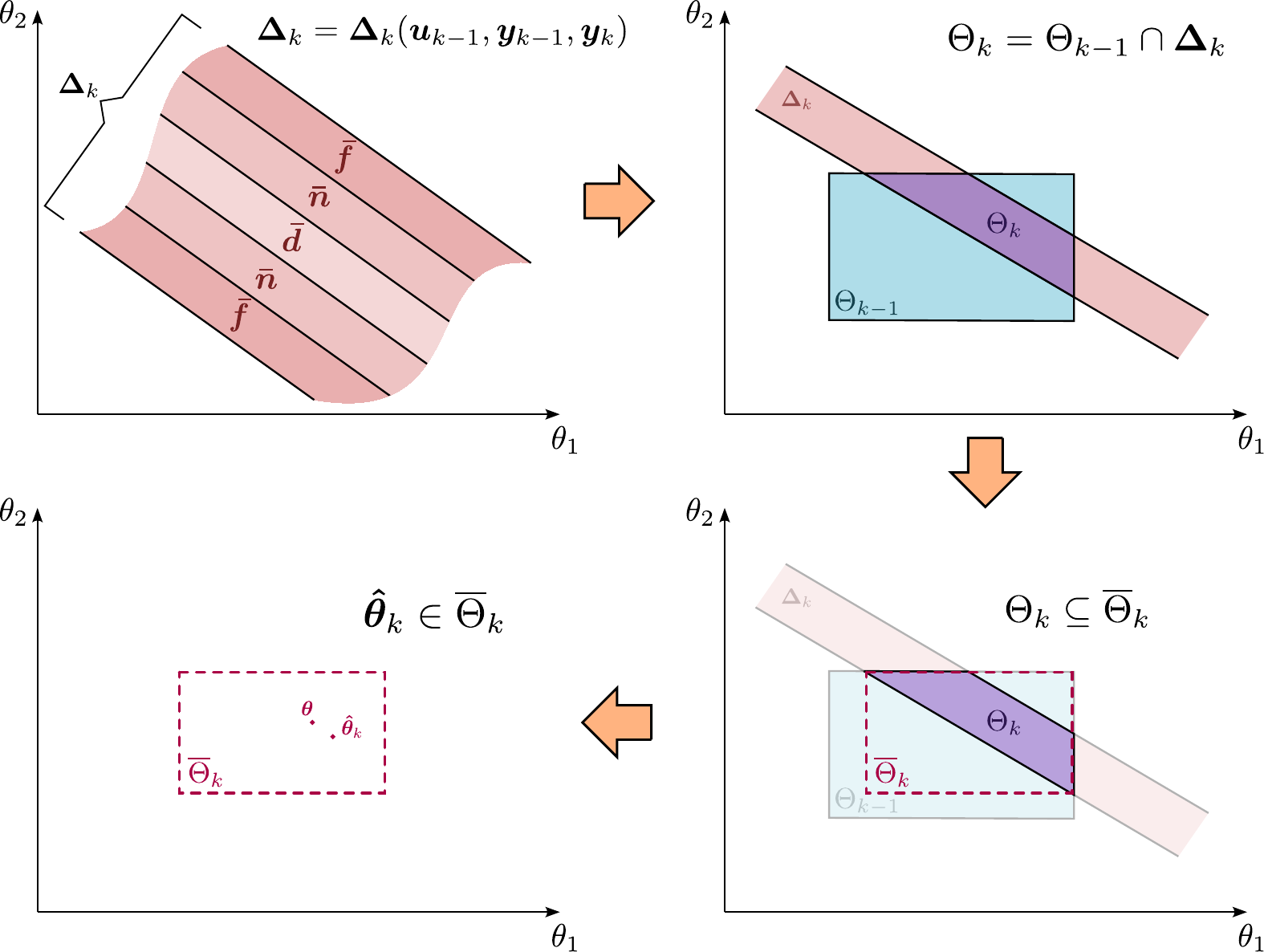}
    \caption{Overview of the different steps in {SME} in a 2-D example: First, the UPS, $\mathrm{\bm{\Delta}}_{k}$, is computed at each time step $k$, based on the latest input-output measurements. The FPS, denoted as $\mathrm{\bm{\Theta}}_k$, is recursively computed based on the intersection of the existing FPS, $\mathrm{\bm{\Theta}}_{k-1}$, and the current update from $\mathrm{\bm{\Delta}}_{k}$. The FPS is then outer-approximated by a simpler polytope, denoted as $\overline{\mathrm{\bm{\Theta}}}_{k}$. Lastly, an estimate, denoted as $\bm{\hat{\theta}}_{k} \in \overline{\mathrm{\bm{\Theta}}}_{k}$, is computed.}
    \label{fig:method_overview}
\end{figure}

The steps involved in SME are illustrated in Figure \ref{fig:method_overview}. First, Section \ref{ss::ups} describes the computation of the Unfalsified Parameter Set (UPS), denoted as $\mathrm{\bm{\Delta}}_{k} \subseteq \mathbb{R}^p$, at each time step $k$, based on the latest input-output measurements (top-left). Subsequently, in Section \ref{ss::fps}, the Feasible Parameter Set (FPS), denoted as $\mathrm{\bm{\Theta}}_k \subseteq \mathbb{R}^p$, is recursively computed based on the intersection of the previous FPS, $\mathrm{\bm{\Theta}}_{k-1}$, and the current update from $\mathrm{\bm{\Delta}}_{k}$ (top-right). The FPS is then outer-approximated by a simpler polytope, denoted as $\overline{\mathrm{\bm{\Theta}}}_{k} \subseteq \mathbb{R}^p$, which is described by a predefined number of maximum directions computed offline according to the required trade-off between accuracy and efficiency (bottom-right). In Section \ref{ss::npe}, an estimate $\bm{\hat{\theta}}_{k} \in \overline{\mathrm{\bm{\Theta}}}_{k}$ is derived, accounting for the quality of the available measurements by incorporating adaptive regularization (bottom-left). Finally, Section \ref{ss::fde} describes the {FDE} method, which relies on the components computed in the preceding sections.

% ======================================================================

\subsection{Unfalsified Parameter Set}\label{ss::ups}
In this section, we introduce a method to compute the Unfalsified Parameter Set (UPS) based on input-output measurements. First, we express the disturbance and noise bounds in polytopic form:
\begin{equation}\label{eq:disturbance_bound_poly}
    |\bm{d}_{k}| \leq \bm{\bar{d}} \Leftrightarrow \bm{d}_{k} \in \mathcal{D} = \{\bm{d}_{k} \in \mathbb{R}^n | \bm{H} \bm{d}_{k} \leq \bm{h_d} \}
\end{equation}
\begin{equation}\label{eq:noise_bound_poly}
    |\bm{n}_{k}| \leq \bm{\bar{n}} \Leftrightarrow \bm{n}_{k} \in \mathcal{N} = \{\bm{n}_{k} \in \mathbb{R}^n | \bm{H} \bm{n}_{k} \leq \bm{h_n} \}
\end{equation}
where $\bm{H} = \begin{bmatrix} \bm{I}_n & -\bm{I}_n \end{bmatrix}^\top \in \mathbb{R}^{2n \times n}$ with $\bm{I}_n \in \mathbb{R}^{n \times n}$ the identity matrix of dimension $n$, $ \bm{h_d} = \begin{bmatrix} \bm{\bar{d}} & \bm{\bar{d}} \end{bmatrix}^\top \in \mathbb{R}^{2n}$ and $ \bm{h_n} = \begin{bmatrix} \bm{\bar{n}} & \bm{\bar{n}} \end{bmatrix}^\top \in \mathbb{R}^{2n}$. Combining equations \eqref{eq:nonlinear_system}  and \eqref{eq:state_measurement} yields:
\begin{equation}\label{eq:nonlinear_system_measurement}
    \bm{y}_{k+1} - \bm{G}(\bm{u}_{k}) \bm{\theta} = \bm{d}_{k} + \bm{n}_{k+1} + \bm{f}(\bm{z}_{k})
\end{equation}
where $\bm{u}_{k}$ and $\bm{y}_{k}$ are known signals (kept on the left-hand side) while $\bm{d}_{k}$ and $\bm{n}_{k}$ are unknown but bounded based on Assumption 1. The complication arises with the remaining term $\bm{f}(\bm{z}_{k})$ which depends on both the known state measurement $\bm{y}_{k}$ and the unknown noise signal $\bm{n}_{k}$ since $\bm{z}_{k}  = \bm{y}_{k} - \bm{n}_{k}$. To handle this term, we will rely on interval analysis \cite{jaulin2001} to compute lower and upper bounds for $\bm{f\left(\cdot\right)}$. To do this, we first need to find an interval for the state $\bm{z}_{k}$ from \eqref{eq:state_measurement} and \eqref{eq:noise_bound} as follows:
\begin{equation}\label{eq:state_bound}
    \underline{\bm{z}}_k = \bm{y}_{k} - \bm{\bar{n}} \leq
    \bm{z}_{k} \leq \bm{y}_{k} + \bm{\bar{n}} = \overline{\bm{z}}_k \Leftrightarrow \bm{z}_k \in [\underline{\bm{z}}_k, \overline{\bm{z}}_k] = [\bm{z}_k]
\end{equation}
which is time-varying and can be updated online with each new state measurement $\bm{y}_{k}$. Usually, an inclusion function $\mathfrak{f}
\left( \cdot \right)$ is found for $\bm{f\left(\cdot\right)}$ based on the state interval \eqref{eq:state_bound} satisfying:
\begin{equation}
    \bm{f}\left( \left[ \bm{z} \right] \right) \subset \mathfrak{f}\left( \cdot \right) 
\end{equation}
where $\bm{f\left( \left[ \bm{z} \right] \right)}$ denotes the minimal inclusion function. Computing $\bm{f\left( \left[ \bm{z} \right] \right)}$ would give the tightest possible bounds, but this requires solving two global, non-convex optimization problems which are prohibitive to solve online. Instead, we can use interval arithmetic which extends math operations and elementary functions to intervals and is much more computationally efficient. Since the expression of $\bm{f\left(\cdot\right)}$ is known, we can compute an interval for the term as:
\begin{equation}\label{eq:autonomous_term_bound}
     [\underline{\bm{f}}_{k}, \overline{\bm{f}}_{k}] = \bm{I}([\underline{\bm{z}}_k, \overline{\bm{z}}_k]) \supseteq \bm{f\left( \left[ \bm{z}_{k} \right] \right)}
\end{equation}
where $\bm{I}\left(\cdot\right)$ denotes an appropriate function to compute intervals via interval arithmetic. In practice, we use C++ BOOST Interval Arithmetic library \cite{bronnimann2003} which can efficiently compute such intervals online. Note that the interval $[\underline{\bm{f}}_{k}, \overline{\bm{f}}_{k}]$ is also time-varying as it implicitly depends on the state measurement $\bm{y}_{k}$. With these known bounds for the system dynamics $\bm{f\left(\cdot\right)}$ we can formulate similar polytopic bounds for the autonomous term in \eqref{eq:nonlinear_system_measurement} similar to those derived before:
\begin{align}\label{eq:autonomous_term_bound_poly}
     \underline{\bm{f}}_{k} \leq \bm{f}(\bm{z}_k) \leq \overline{\bm{f}}_{k} 
     &\Leftrightarrow \nonumber \\
     \bm{f}(\bm{z}_k) \in \mathcal{F}_{k} 
     & = \{\bm{f}(\bm{z}_k) \in \mathbb{R}^n \, | \, \bm{H} \bm{f}(\bm{z}_k) \leq \bm{h_{f}}(\bm{y}_{k}) \}
\end{align}
with $ \bm{h_{f}}(\bm{y}_{k}) = \begin{bmatrix} \overline{\bm{f}}_{k} & -\underline{\bm{f}}_{k} \end{bmatrix}^\top \in \mathbb{R}^{2n}$. The key observation is that the polytopic inequalities in \eqref{eq:disturbance_bound_poly}, \eqref{eq:noise_bound_poly} and \eqref{eq:autonomous_term_bound_poly} are constructed such that all the unknown signals are multiplied from the left with the same matrix $\bm{H}$. Additionally, from \eqref{eq:noise_bound} we know that $|\bm{n}_{k}| \leq \bm{\bar{n}}, \, \forall k$ and thus from \eqref{eq:noise_bound_poly} we can also deduce that $\bm{H} \bm{n}_{k+1} \leq \bm{h_n}$. We can then sum these inequalities and factor out $\bm{H}$ to get:
\begin{equation}\label{eq:polytopic_inequalities_summation}
    \bm{H}(\bm{d}_{k}+\bm{n}_{k+1}+\bm{f}(\bm{z}_{k})) \leq \bm{h_d} + \bm{h_n} + \bm{h_{f}}(\bm{y}_{k})
\end{equation}
Notice that the right-hand side term in \eqref{eq:nonlinear_system_measurement} appears in the left-hand side of \eqref{eq:polytopic_inequalities_summation}. After substituting \eqref{eq:nonlinear_system_measurement} to \eqref{eq:polytopic_inequalities_summation} and rearranging the terms we get:
\begin{equation}\label{eq:unfalsified_parameter_set_before_shift}
    - \bm{H}\bm{G}(\bm{u}_{k}) \bm{\theta} \leq \bm{h_d} + \bm{h_n} + \bm{h_{f}}(\bm{y}_{k}) - \bm{H}\bm{y}_{k+1}
\end{equation}

\begin{figure}[tbh!]
    \centering
    \includegraphics[width=\columnwidth]{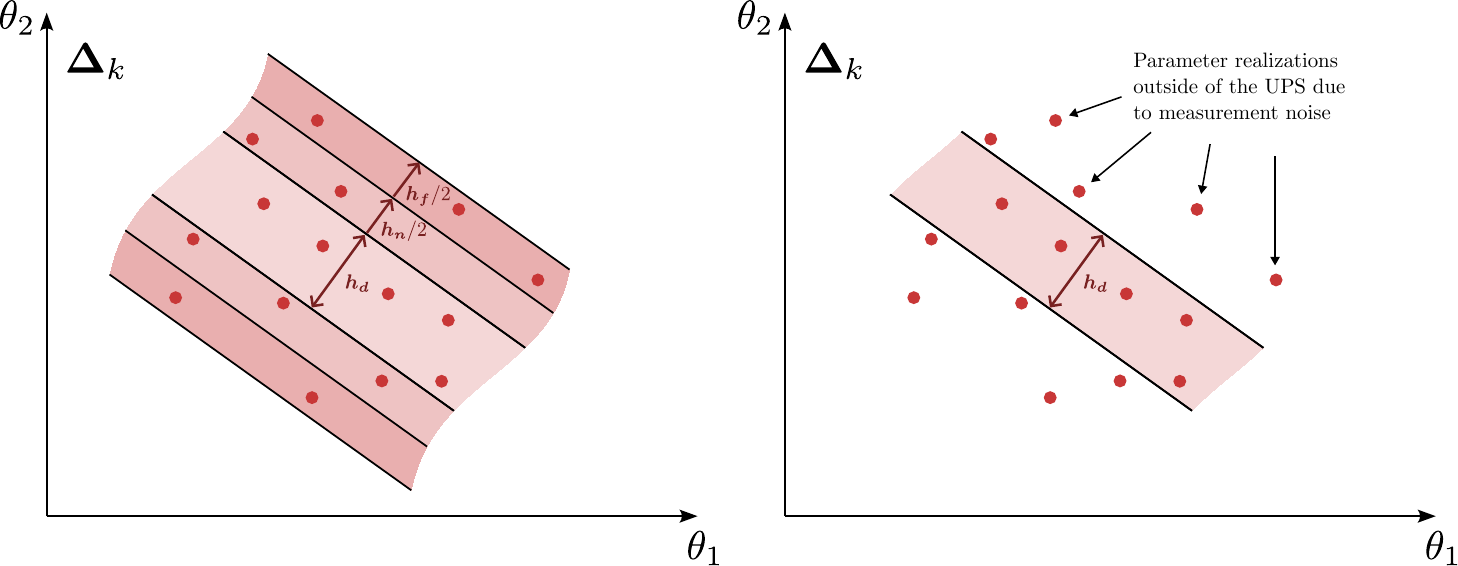}
    \caption{On the right is our formulation of the UPS that additionally accounts for measurement noise and thus all possible parameter realizations are guaranteed to lie inside the UPS. In contrast, if the noise is not accounted for, several realizations will be outside the UPS.}
    \label{fig:ups}
\end{figure}

If we now shift the timestep one step backward, we can derive the UPS as:
\begin{equation}\label{eq:unfalsified_parameter_set}
    \bm{\Delta}_{k} = \{ \bm{\theta} \in \mathbb{R}^p \mid \; - \bm{H}\bm{G}(\bm{u}_{k-1}) \bm{\theta}
    \leq \bm{h_d} + \bm{h_n} + \bm{h_f}(\bm{y}_{k-1}) - \bm{H}\bm{y}_{k} \} 
\end{equation}
which can be computed at each time step $k$ based on the input-output measurement set $\{\bm{u}_{k-1}, \bm{y}_{k-1}, \bm{y}_{k}\}$. Notice that \eqref{eq:unfalsified_parameter_set} is a similar expression to the ones found in \cite{lu2019} and \cite{didier2021} with the difference that here we have additional bounding terms on the right-hand side of the inequality in \eqref{eq:unfalsified_parameter_set} to account for measurement noise: Term $\bm{h_n}(\bm{\bar{n}})$ directly sums the measurement bounds while $\bm{h_{f}}(\bm{y}_{k-1},\bm{\bar{n}})$ is a time-varying term that sums the noise bounds implicitly after they are mapped through the nonlinear autonomous term $\bm{f}(\bm{z}_{k}) = \bm{f}(\bm{y}_{k}, \bm{n}_{k})$ via interval arithmetic. This is illustrated schematically in Figure \ref{fig:ups} in comparison with not accounting for the measurement noise. The UPS is therefore dilated due to the measurement noise as in \cite{didier2021} but its dilation is generalized here to nonlinear systems that are linear to the parameters by employing interval analysis. Notice that in this formulation, even if the system is nonlinear, the assumption that the system is linear to the parameters maintains the same polytopic description for the UPS as a $p$-dimensional slab in the parameter space which is advantageous for the computation of the FPS, $\mathrm{\bm{\Theta}}_k$, as explained in the next section.

% ======================================================================

\subsection{Feasible Parameter Set}\label{ss::fps}
\begin{figure}[tbh!]
    \centering
    \includegraphics[width=\columnwidth]{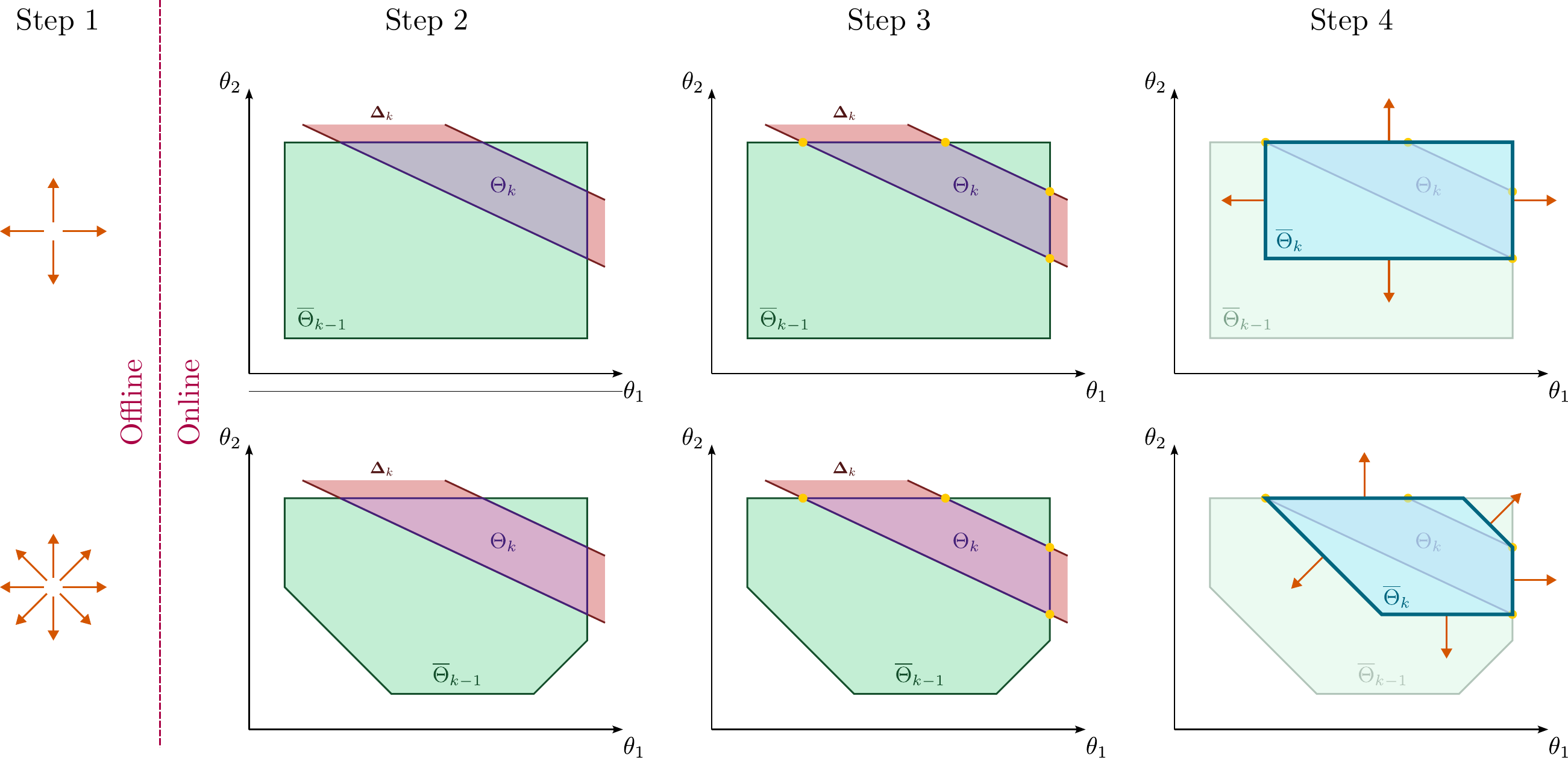}
    \caption{The steps to compute the FPS. This starts with computing the predefined directions offline (Step 1). Then the process continues with computing the new FPS from the current UPS (Step 2), computing the vertices of the new FPS (Step 3), and lastly outer-approximating the new FPS based on the predefined directions (Step 4). The process starts again from Step 2, starting from the new outer approximation. Different choices of predefined directions will result in a different outer approximation of the FPS (top and bottom rows).}
    \label{fig:fps}
\end{figure}

In order to compute the Feasible Parameter Set (FPS), $\mathrm{\bm{\Theta}}_k$, we begin with expressing some initial parameter bounds in polytopic form:
\begin{equation}\label{eq:parameter_bound_poly}
    \bm{\theta} \in \mathrm{\bm{\Theta}}_0 = \{\bm{\theta} \in \mathbb{R}^p | \bm{H}_{\bm{\theta}_0} \bm{\theta} \leq \bm{h}_{\bm{\theta}_0} \}
\end{equation}
where $\bm{H}_{\bm{\theta}_0} = \begin{bmatrix} \bm{I}_p & -\bm{I}_p \end{bmatrix}^\top \in \mathbb{R}^{2p \times p}$ with $\bm{I}_p \in \mathbb{R}^{p \times p}$ the identity matrix of dimension $p$, and $\bm{h_{\theta_{0}}} = \begin{bmatrix} \bm{\underline{\theta}} & \,\bm{\overline{\theta}} \end{bmatrix}^\top$
has known, lower and upper parameter bounds. The parameter set $\mathrm{\bm{\Theta}}_0$ is defined as the initial $p$-dimensional hypercube which describes the range of values of the parameters of interest. The FPS is recursively updated using the UPS, $\bm{\Delta}_{k}$, from \eqref{eq:unfalsified_parameter_set} at each time step starting from the initial FPS, $\mathrm{\bm{\Theta}}_{0}$, as:
\begin{equation}\label{eq:feasible_parameter_set}
    \mathrm{\bm{\Theta}}_k = \mathrm{\bm{\Theta}}_{k-1} \cap \bm{\Delta}_{k}, \quad \forall k=1,2,...,N
\end{equation}
If we describe the FPS in polytopic form:
\begin{equation}\label{eq:feasible_parameter_set_poly}
    \mathrm{\bm{\Theta}}_k = \{\bm{\theta} \in \mathbb{R}^p | \bm{H}_{\bm{\theta}_k} \bm{\theta} \leq \bm{h}_{\bm{\theta}_k} \}
\end{equation} 
and rewrite \eqref{eq:unfalsified_parameter_set} with a simplified notation as:
\begin{equation}\label{eq:unfalsified_parameter_set_simple}
    \bm{\Delta}_{k} = \{ \bm{\theta} \in \mathbb{R}^p \mid \; \bm{H}_{\bm{\Delta}_{k}} \bm{\theta} \leq \bm{h}_{\bm{\Delta}_{k}} \}
\end{equation}
with $\bm{H}_{\bm{\Delta}_{k}}=- \bm{H}\bm{G}(\bm{u}_{k-1})$ and $\bm{h}_{\bm{\Delta}_{k}}=\bm{h_d} + \bm{h_n} + \bm{h_{f}}(\bm{y}_{k-1}) - \bm{H}\bm{y}_{k}$ then the intersection of sets described in \eqref{eq:feasible_parameter_set} is the concatenation of the inequalities that characterize the FPS in \eqref{eq:feasible_parameter_set_poly} and the UPS in \eqref{eq:unfalsified_parameter_set_simple}:
\begin{equation}\label{eq:feasible_param_set_concatenate}
    \bm{H}_{\bm{\theta}_k} = 
    \begin{bmatrix}
        \bm{H}_{\bm{\theta}_{k-1}} \\
        \bm{H}_{\bm{\Delta}_{k}}
    \end{bmatrix} , \quad
    \bm{h}_{\bm{\theta}_k} = 
    \begin{bmatrix}
        \bm{h}_{\bm{\theta}_{k-1}} \\
        \bm{h}_{\bm{\Delta}_{k}}
    \end{bmatrix}
\end{equation}
This process works on the condition that the latest set of measurements is informative enough to ensure that the newly introduced inequalities are not redundant. However, continuously concatenating the inequalities, expressed as $\bm{H}_{\bm{\theta}_k} = \begin{bmatrix} \bm{H}_{\bm{\theta}_{k-N}} & ... & \bm{H}_{\bm{\theta}_{k-1}} & \bm{H}_{\bm{\theta}_k} \end{bmatrix}^\top \in \mathbb{R}^{2Np \times p}$ and $\bm{h}_{\bm{\theta}_k} = \begin{bmatrix} \bm{h}_{\bm{\theta}_{k-N}} & ... & \bm{h}_{\bm{\theta}_{k-1}} & \bm{h}_{\bm{\theta}_k} \end{bmatrix}^\top \in \mathbb{R}^{2Np}$ quickly becomes impractical since the size of $\bm{H}_{\bm{\theta}_k}$ and $\bm{h}_{\bm{\theta}_k}$ grows unbounded as $N \rightarrow \infty$. In \cite{lu2019} this is handled by outer-approximating the FPS with hypercubes encompassing the derived polytope at each timestep $k$. In \cite{lorenzen2019} this is generalized by using predefined normal directions of the facets of a polytope that bounds the estimated parameter set and then solving an optimization problem to ``tighten" the polytope around the FPS. We propose a method that can efficiently provide an outer bound of the FPS with the accuracy of the bound adjustable depending on the balance between required precision and available computational resources. The computation of the FPS is illustrated in Figure \ref{fig:fps} and consists of the following steps:
\begin{enumerate}
    \item Recursively compute a set of predefined, normalized directions, $\mathcal{E} = \{\bm{e}_1, \bm{e}_2, ... \}$, normal to the faces of the outer-approximating polytope. These directions are based on the number of parameters $p$ and a user-defined accuracy iterator $\phi$ from Algorithm \ref{alg:predefined_directions}. This is computed offline.
    \item Compute the new FPS $\mathrm{\bm{\Theta}}_k = \overline{\mathrm{\bm{\Theta}}}_{k-1} \cap \bm{\Delta}_{k}$ based on the outer-approximation of the previous time step and the new UPS computed at time $k$ starting with $\overline{\mathrm{\bm{\Theta}}}_{0} = {\mathrm{\bm{\Theta}}}_{0}$.
    \item Compute the set of vertices $\mathcal{V}_k = \{\bm{v}^1_k, \bm{v}^2_k, ... \}$ of the convex polytope $\mathrm{\bm{\Theta}}_k$.
    \item Compute the outer approximation $\overline{\mathrm{\bm{\Theta}}}_{k}$ of the convex polytope $\mathrm{\bm{\Theta}}_k$ based on the set of vertices $\mathcal{V}_k$ and the set of predefined directions $\mathcal{E}$ from Algorithm \ref{alg:outer_approximation}.
    \item Go back to Step 2.
\end{enumerate}
We propose an algorithm that systematically generates predefined directions for the faces of the approximation polytope offline, with arbitrarily high complexity, and for any number of parameters \( p \). The generation of predefined directions \( \mathcal{E} \) begins with a simple \( p \)-dimensional hypercube. The key idea is to ``bisect" each edge formed by the intersection of adjacent faces and create a new face with a normal vector that points symmetrically between the normal directions of the intersecting faces. This process can be repeated recursively with a user-defined number of recursions, \( \phi \in \mathbb{N} \). As \( \phi \rightarrow \infty \), the predefined normal directions begin to approximate a \( p \)-dimensional sphere, allowing the polytope to closely approximate any convex shape, at the expense of increased computational complexity. The algorithm for generating these predefined directions is outlined in Algorithm \ref{alg:predefined_directions} and is intended to run offline.
\algrenewcommand\algorithmicrequire{\textbf{Input:}}
\algrenewcommand\algorithmicensure{\textbf{Output:}}
\begin{algorithm}
\caption{Generate Predefined Directions (Offline)}\label{alg:predefined_directions}
\begin{algorithmic}[1]
    \Require $p$ , $\phi$
    \Ensure $\mathcal{E}$
    \State $\mathcal{E} \gets \{\pm \bm{e}_i : i = 1, \dots, p\}$
    \For{$i \gets 1$ \textbf{to} $\phi$}
        \State $\mathcal{E}^{\text{new}} \gets \emptyset$
        \For{$j \gets 1$ \textbf{to} $p$}
            \For{each combination of $j$ elements in $\mathcal{E}$}
                \State $\bm{e}^{\text{new}} \gets$ sum of the combination
                \If{$\|\bm{e}^{\text{new}}\| \neq 0$}
                    \State $\bm{e}^{\text{new}} \gets \frac{\bm{e}^{\text{new}}}{\|\bm{e}^{\text{new}}\|}$ 
                    \State $\mathcal{E}^{\text{new}} \gets \mathcal{E}^{\text{new}} \cup \{\bm{e}^{\text{new}}\}$
                \EndIf
            \EndFor
        \EndFor
        \State $\mathcal{E} \gets \mathcal{E}^{\text{new}}$
    \EndFor
\end{algorithmic}
\end{algorithm}

The set of vertices $\mathcal{V}_k$ of the convex polytope $\mathrm{\bm{\Theta}}_k$ is computed by first removing redundant inequalities with a set of Linear Programs ({LP}s) solved with \cite{glpk} and then by computing the solution of all possible combinations of the linear algebraic equations that describe the non-redundant inequalities. The solutions that also satisfy the inequalities are stored as the vertices of the convex polytope. Having the set predefined directions $\mathcal{E}$ from Algorithm \ref{alg:predefined_directions} and the set of vertices $\mathcal{V}_k$ of the convex polytope $\mathrm{\bm{\Theta}}_k$, we need to outer-approximate the FPS given in \eqref{eq:feasible_parameter_set} at each timestep $k$ such that $\mathrm{\bm{\Theta}}_k \subseteq \overline{\mathrm{\bm{\Theta}}}_k$. In contrast to \cite{lorenzen2019} where an optimization problem is solved, we use linear algebra to compute the extremum vertices of the polytopic set $\mathrm{\bm{\Theta}}_k$ along the directions $\mathcal{E}$. This process is described in Algorithm \ref{alg:outer_approximation}.

\algrenewcommand\algorithmicrequire{\textbf{Input:}}
\algrenewcommand\algorithmicensure{\textbf{Output:}}
\begin{algorithm}
\caption{Outer-approximate Convex Polytope}
\label{alg:outer_approximation}
\small
\begin{algorithmic}[1] % Numbering lines
    \Require $\mathcal{E}$, $\mathcal{V}_k$
    \Ensure $\overline{\mathrm{\bm{\Theta}}}_{k}$ as the pair $\langle \bm{\overline{H}}_{\bm{\theta}_k}, \bm{\overline{h}}_{\bm{\theta}_k} \rangle$
    \For{$\bm{e}_i \in \mathcal{E}$ }
        \State $\Pi \gets -\infty$
        \For{$\bm{v}^j_k \in \mathcal{V}_k$ }
            \If{$\bm{e}_i^{\top} \bm{v}^j_k > \Pi$}
                \State $\Pi \gets \bm{e}_i^{\top} \bm{v}^j_k$
                \State $\overline{\bm{v}}^i_k \gets \bm{v}^j_k$
            \EndIf
        \EndFor
         \State $\bm{\overline{H}}_{\bm{\theta}_k}(i,:) \gets {\overline{\bm{v}}^i_k}^{\top} $
        \State $\bm{\overline{h}}_{\bm{\theta}_k}(i) \gets \bm{e}_i^{\top}\overline{\bm{v}}^i_k $
    \EndFor

    \State $\langle \bm{\overline{H}}_{\bm{\theta}_k}, \bm{\overline{h}}_{\bm{\theta}_k} \rangle \gets \text{remove\_redundant\_constraints}(\bm{\overline{H}}_{\bm{\theta}_k}, \bm{\overline{h}}_{\bm{\theta}_k})$
\end{algorithmic}
\end{algorithm}

% =======================================================================

\subsection{Parameter Estimate}\label{ss::npe}
The final step is to derive an estimate $\bm{\hat{\theta}}_k$ for the unknown parameter $\bm{\theta}_k$ that belongs to the derived parameter set $\overline{\mathrm{\bm{\Theta}}}_k$. We can exploit again here the fact that the system is linear to the parameters of interest and use the following equation:
\begin{equation}\label{eq:regression_equation_1}
    \bm{G}(\bm{u}_{k-1}) \bm{\theta} = \bm{y}_{k} - \bm{f}(\bm{y}_{k-1})
\end{equation}
which is a linear algebraic equation to the unknown parameter $\bm{\theta}_k$ and where the disturbance and the noise are included in the measurement. If we concatenate \eqref{eq:regression_equation_1} for the last $N$ measurements, to leverage more data, we can get a regression equation:
\begin{equation}\label{eq:regression_equation_2}
\underbrace{\begin{bmatrix}
    \bm{G}(\bm{u}_{k-1-N}) \\
    ... \\
     \bm{G}(\bm{u}_{k-1}) \\
\end{bmatrix}}_{\bm{\Phi}} \bm{\theta} =
\underbrace{\begin{bmatrix}
    \bm{y}_{k-N} - \bm{f}(\bm{y}_{k-1-N}) \\
    ... \\
    \bm{y}_{k} - \bm{f}(\bm{y}_{k-1}) \\
\end{bmatrix}}_{\bm{\xi}}
\end{equation}
where $\bm{\theta}$ here is the \textit{regressand}, $\bm{\Phi}$ is the \textit{regressor} and $\bm{\xi}$ the \textit{observation}.  The solution to the unconstrained classical Least Squares Problem ({LSP}) has a well-known form using the expression of the Moore-Penrose pseudo-inverse. However, here we want $\bm{\hat{\theta}}_k \in \overline{\mathrm{\bm{\Theta}}}_k$ so a closed-form solution cannot be used and instead a Quadratic Program ({QP}) needs to be solved online considering the linear inequality constraints introduced from $\overline{\mathrm{\bm{\Theta}}}_k$. Furthermore, the rank of the regressor matrix $\bm{\Phi}$—which depends on input measurements—directly impacts the solvability and quality of the solution to the parameter estimation problem as the regressor is not always guaranteed to be full rank. A rank-deficient matrix implies that not all parameters in $\bm{\theta}$ are independently estimable from the given inputs, leading to either non-unique solutions or inaccurate estimates. To address these issues, we formulate the following {QP} with generalized Tikhonov regularization:
\begin{equation}\label{eq:quadratic_program_estimate}
\begin{aligned}
& \underset{\bm{\theta}}{\text{min}}
& & \bm{\theta}^{\top} \bm{P} \bm{\theta} + \bm{q}^{\top} \bm{\theta} \\
& \text{s.t.:}
& & \bm{\overline{H}}_{\bm{\theta}_k} \bm{\theta} \leq \bm{\overline{h}}_{\bm{\theta}_k}, \\
&&& \bm{\theta}^0 = \bm{\theta}^c_k.
\end{aligned}
\end{equation}
with $\bm{P} = \bm{\Phi}^{\top}\bm{\Phi} + \bm{\Lambda}$, $\bm{q}=-2(\bm{\xi}^{\top}\bm{\Phi}+\tilde{\bm{\theta}}^{\top} \bm{\Lambda})$, $\bm{\Lambda}$ the regularization matrix, $\tilde{\bm{\theta}}$ the regularization value of $\bm{\theta}$, $\bm{\theta}^c_k$ the vertex centroid of polytope $\overline{\mathrm{\bm{\Theta}}}_k$, given as $\frac{1}{N_v} \sum_{i=1}^{N_v} \overline{\bm{v}}_i$, and $\bm{\theta}^0$ the initial guess for the solution of the {QP} problem. Since the regressor matrix $\bm{\Phi}$ depends on a window of input measurements and has a time-varying rank condition, regularization should only be significant when the matrix approaches rank deficiency and should be negligible otherwise. Therefore, we introduce an adaptive regularization parameter $\bm{\Lambda}$ as an exponential decay function of the rank condition of the regressor matrix:
\begin{equation}
    \bm{\Lambda} = \bar{\bm{\Lambda}} e^{-\alpha \bm{\Sigma}_p}
\end{equation} 
where $\bar{\bm{\Lambda}}$ is the maximum value of the regularization parameter matrix, $\alpha$ is a parameter to tune the decay rate of the function, and $\bm{\Sigma}_p$ is the diagonal singular value matrix that results from the compact Singular Value Decomposition ({SVD}) of $\bm{\Phi}$. Because of the efficiency of {QP} solvers along with the limited problem dimension, (bounded dimensions of $\bm{\overline{H}}_{\bm{\theta}_k} \bm{\theta} \leq \bm{\overline{h}}_{\bm{\theta}_k}$ and fixed size of parameter vector $p$) the {QP} \eqref{eq:quadratic_program_estimate} can be solved swiftly online with \cite{osqp}.

%=======================================================================

\subsection{Fault Decision Logic}\label{ss::fde}
\begin{figure}
    \centering
    \includegraphics[width=\columnwidth]{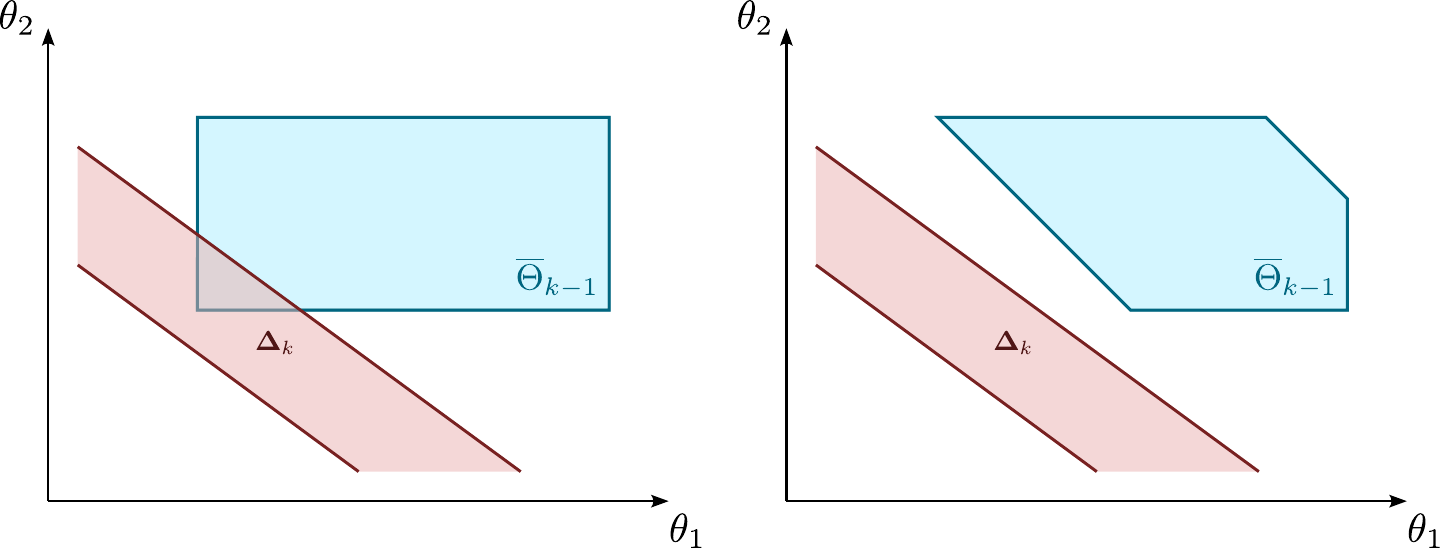}
    \caption{Schematic representation of fault detection using the inverse test on the FPS. In this example, the UPS arises from a measurement corrupted by a fault. In the left figure, the outer approximation of the FPS is more conservative, preventing fault detection, as $\mathrm{\overline{\boldsymbol{\Theta}}}_{k-1} \cap \boldsymbol{\Delta}_{k} \neq \emptyset$. In the right figure, due to a tighter outer approximation of the FPS, the same measurement results in $\mathrm{\overline{\boldsymbol{\Theta}}}_{k-1} \cap \boldsymbol{\Delta}_{k} = \emptyset$, successfully revealing the fault.}
    \label{fig:fault_detection}
\end{figure}

The different components of SME outlined in previous sections are utilized to perform FD, as summarized in Algorithm \ref{alg:sme_fault_diagnosis}. As new input-output measurements are obtained, the UPS and FPS are updated, alongside the outer approximation of the FPS and the parameter estimate. If the input is bound exploring, we can obtain minimal uncertainty. Thus, under healthy conditions, the FPS typically converges to a ``healthy set" of fault parameter values around the nominal value of a healthy parameter while taking into account the disturbance and noise bounds described in Assumption 1. If, at any timestep \( k_F \), a fault occurs, the newly computed UPS will likely no longer intersect with the current FPS, depending on the fault's severity (see Figure \ref{fig:fault_detection}). This implies that the most recent data provides a set of parameters that do not belong to the ``healthy FPS", leading to the conclusion that a fault has been detected. 

\textit{Fault Detection Logic:} If $\overline{\mathrm{\boldsymbol{\Theta}}}_{k-1} \cap \boldsymbol{\Delta}_k = \emptyset$, then a fault is guaranteed to be detected.

% Definition

A fault is detected at the first timestep \( k_D \) when the following condition occurs:
\begin{equation}\label{eq:k_D}
    k_D = \min \{ k \mid \overline{\mathrm{\boldsymbol{\Theta}}}_{k-1} \cap \boldsymbol{\Delta}_k = \emptyset, \; k > k_F \}
\end{equation}
where $k_F$ indicates that timestep a fault has occurred. To isolate the fault, we follow a similar approach as for the detection, but now we need to check the projection of the FPS on the different principle axes of the parameter space first. The projection of the FPS on the principle axes of the parameter space is given by:
\begin{equation}\label{eq:FPS_projection}
    \text{Proj}_{\theta_i}(\mathrm{\boldsymbol{\Theta}}_k) = \left[\min_{\boldsymbol{v} \in \mathcal{V}} (e_{\theta_i}^\top \boldsymbol{v}), \max_{\boldsymbol{v} \in \mathcal{V}} (e_{\theta_i}^\top \boldsymbol{v})\right]
\end{equation}
where \( e_{\theta_i}, \; i = 1, 2, \dots, p \) denotes the unit vectors of the orthonormal basis of the parameter space, and \( \boldsymbol{v} \in \mathcal{V} \) represents the vertices of the FPS. Since the FPS is a convex set by construction, this one-dimensional projection results in an interval. We can then compare these intervals before and after fault detection.

\textit{Fault Isolation Logic:} If $\text{Proj}_{\theta_i}(\mathrm{\boldsymbol{\Theta}}_{k}) \cap \text{Proj}_{\theta_i}(\mathrm{\boldsymbol{\Theta}}_{k_D-1}) = \emptyset$, then $\theta_i$ is guaranteed to be faulty. 

A fault is isolated at the first timestep $k^i_I, \; i = 1,2,\dots,p$ when the following condition occurs:
\begin{equation}\label{eq:k_I}
    k^i_I = \min \{ k \mid \text{Proj}_{\theta_i}(\mathrm{\boldsymbol{\Theta}}_{k}) \cap \text{Proj}_{\theta_i}(\mathrm{\boldsymbol{\Theta}}_{k_D-1}) = \emptyset, \; k > k_D \}
\end{equation}
Following fault detection and isolation, the containers are reinitialized, and the parameter set begins to converge toward a ``faulty" FPS, accompanied by a new estimate for the fault parameters.
\begin{algorithm}
\caption{SME-based Fault Diagnosis}
\label{alg:sme_fault_diagnosis}
\small
\begin{algorithmic}[1] % Numbering lines
    \Require $\boldsymbol{\bar{d}}$, $\boldsymbol{\bar{n}}$, $\mathrm{\boldsymbol{\Theta}}_0$, $\boldsymbol{f}(\cdot)$, $\boldsymbol{G}(\cdot)$, $\phi$
    \Ensure $\overline{\mathrm{\boldsymbol{\Theta}}}_{k}$, $\boldsymbol{\hat{\theta}}_k \in \overline{\mathrm{\boldsymbol{\Theta}}}_k$
    \State $\textrm{Compute predefined directions, } \mathcal{E} \textrm{, from Algorithm \ref{alg:predefined_directions} } $
    \For{$k = 1,2,\dots$}
        \State $\textrm{Get input-output data } \{\boldsymbol{y}_{k}, \boldsymbol{y}_{k-1}, \boldsymbol{u}_{k} \}$
        \State $\textrm{Compute UPS, } \boldsymbol{\Delta}_{k} \textrm{, from \eqref{eq:unfalsified_parameter_set_simple} } $
        \State $\textrm{Compute FPS, } \mathrm{\boldsymbol{\Theta}}_k \textrm{, from \eqref{eq:feasible_param_set_concatenate} } $
        \State $\textrm{Compute projection of } \mathrm{\boldsymbol{\Theta}}_k \textrm{, from \eqref{eq:FPS_projection} }$ 
        \State $\textrm{Check feasibility of FPS, } \mathrm{\boldsymbol{\Theta}}_k,  \textrm{ (with {LP}s)}$
        \If{$\mathrm{\boldsymbol{\Theta}}_k = \emptyset$}
            \State $\textrm{Fault detected}$
            \State $k_D=k$
            \For{$i = 1,2,\dots,p$} 
                \If {$\text{Proj}_{\theta_i}(\mathrm{\boldsymbol{\Theta}}_{k_D}) \cap \text{Proj}_{\theta_i}(\mathrm{\boldsymbol{\Theta}}_{k_D-1}) = \emptyset $}
                    \State $\textrm{Fault isolated at $\theta_i$}$
                    \State $k^i_I=k$
                \EndIf
            \EndFor   
            \State $\mathrm{\boldsymbol{\Theta}}_k = \mathrm{\boldsymbol{\Theta}}_0$
            \State $\boldsymbol{\Phi}=[]$, $\boldsymbol{\xi}=[]$
        \EndIf 
        \State $\textrm{Compute outer-approximation}, \overline{\mathrm{\boldsymbol{\Theta}}}_k \textrm{, from Algorithm \ref{alg:outer_approximation} } $
        \State $\textrm{Compute vertex centroid as } \boldsymbol{\theta}^c_k = \frac{1}{N_v} \sum_{i=1}^{N_v} \overline{\boldsymbol{v}}_i$
        \State $\textrm{Compute estimate }, \boldsymbol{\hat{\theta}}_k \in \overline{\mathrm{\boldsymbol{\Theta}}}_k \textrm{, from \eqref{eq:quadratic_program_estimate} } $
    \EndFor
\end{algorithmic}
\end{algorithm}

\section{Case Study: Autonomous Surface Vehicle}\label{s:results}
In this case study we consider the 3-DoF model of an ASV in planar motion, equipped with sensors, actuators, and the trajectory optimizer from \cite{tsolakis2024a} for path-following and collision avoidance. The ASV dynamics follow the maneuvering model in \cite{fossen2011}. Its state includes position $\bm{p} = (x,y)^\top$, orientation $\psi$, longitudinal velocity $u$, lateral velocity $v$, and yaw rate $r$, expressed in the body-fixed frame and denoted as $\bm{z} = (x,y,\psi,u,v,r)^\top \in \mathcal{Z} \subset \mathbb{R}^6$, while the control input $\bm{u} = (\tau_l,\tau_r,\tau_b,\alpha_l,\alpha_r)^\top \in \mathcal{U} \subset \mathbb{R}^5$ represents the actions of two azimuth thrusters and a bow thruster. Specifically, $\tau_l$ and $\tau_r$ are the thrusts, and $\alpha_l$ and $\alpha_r$ are the azimuth angles of the left and right thrusters, while $\tau_b$ represents the thrust of the bow thruster. Environmental disturbances from wind and waves are denoted as $\bm{\tau_d}$. The system's evolution is governed by the following continuous, nonlinear system:
{\small
\renewcommand{\arraystretch}{0.1} % Adjusts the row spacing in matrices
\begin{equation}\label{eq:zdot}
\dot{\bm{z}} = \underbrace{
\begin{bmatrix}
    \bm{0}_{3\times3} & \bm{R}(\bm{z})\\
    \bm{0}_{3\times3} & -\bm{M}^{-1} (\bm{C}(\bm{z})+\bm{D}(\bm{z}))
\end{bmatrix} 
\bm{z}}_{\bm{f}(\bm{z})}
+ \underbrace{
\begin{bmatrix}
    \bm{0}_{3\times3}\\
    \bm{M}^{-1}
\end{bmatrix} 
\bm{\tau}(\bm{u})}_{\bm{{g}}(\bm{u})}
+ \underbrace{
\begin{bmatrix}
    \bm{0}_{3\times3}\\
    \bm{M}^{-1}
\end{bmatrix} 
\bm{\tau_d}}_{\bm{d}}
\end{equation}}
Here, $\bm{R}(\bm{z})$ is the rotation matrix, $\bm{M}$ is the mass matrix, $\bm{C}(\bm{z})$ is the Coriolis and centripetal matrix and $\bm{D}(\bm{z})$ the damping matrix. The generalized force vector acting on the ASV is denoted by $\bm{\tau}$, and $w_{lr}$, $l_{lr}$, and $l_b$ are length parameters that define the thruster configuration. The thrust generated by the actuators under healthy conditions is represented by $\bm{\tau}(\bm{u})$, and actuator limitations are also taken into account. For fault modeling, we introduce actuator faults represented by $\bm{\theta} = \left( \theta_l, \theta_r, \theta_b \right)^{\top}$, where $\theta_l$, $\theta_r$, and $\theta_b$ denote the loss of effectiveness (LoE) in the left, right, and bow thruster, respectively expressed in polytopic form:
\begin{equation}\label{eq:fault_poly}
    \begin{bmatrix} \bm{I}_p & -\bm{I}_p \end{bmatrix} \bm{\theta} \leq  \begin{bmatrix} \bm{1} & \bm{0} \end{bmatrix}
\end{equation}
The input map can be written linearly to the parameters as:
{\small
\setlength{\arraycolsep}{3pt} % Adjust column spacing as needed
\begin{equation}
    \bm{{g}}(\bm{u}) = \underbrace{
    \begin{bmatrix}
    \bm{0}_{3\times3}\\
    \bm{M}^{-1}
    \end{bmatrix} 
    \begin{bmatrix}
        \tau_l \cos{\alpha_l} & \tau_r \cos{\alpha_r} & 0 \\
        \tau_l \sin{\alpha_l} & \tau_r \sin{\alpha_r} & 0 \\
        \begin{array}{c} -w_{lr} \tau_l \cos{\alpha_l} \\ \quad -l_{lr} \tau_l \sin{\alpha_l} \end{array} & 
        \begin{array}{c} w_{lr} \tau_r \cos{\alpha_r} \\ \quad -l_{lr} \tau_r \sin{\alpha_r} \end{array} & 
        l_b \tau_b \\
    \end{bmatrix}}_{\bm{{G}}(\bm{u})}
    \underbrace{\begin{pmatrix}
        \theta_l \\
        \theta_r \\
        \theta_b
    \end{pmatrix}}_{\bm{\theta}}
\end{equation}}
We assume full-state measurements with additive measurement noise. The disturbances and measurement noise follow a uniform distribution with known bounds chosen as $\bm{\bar{d}}=(0.02,0.03,0.003,0.02,0.03,0.01)^\top$ and $\bm{\bar{n}}=(0.01,0.01,0.001,0.007,0.005,0.012)^\top$ respectively. The dynamics in \eqref{eq:zdot} are discretized using Runge-Kutta for numerical implementation. Our framework is implemented in ROS: the controller and FD module in C++ and the simulator in Python. The algorithm runs in an Ubuntu machine with an Intel i7 CPU@1.8GHz and 16GB of RAM. 

\begin{figure}[tbh!]
  \centering
  \includegraphics[width=\linewidth]{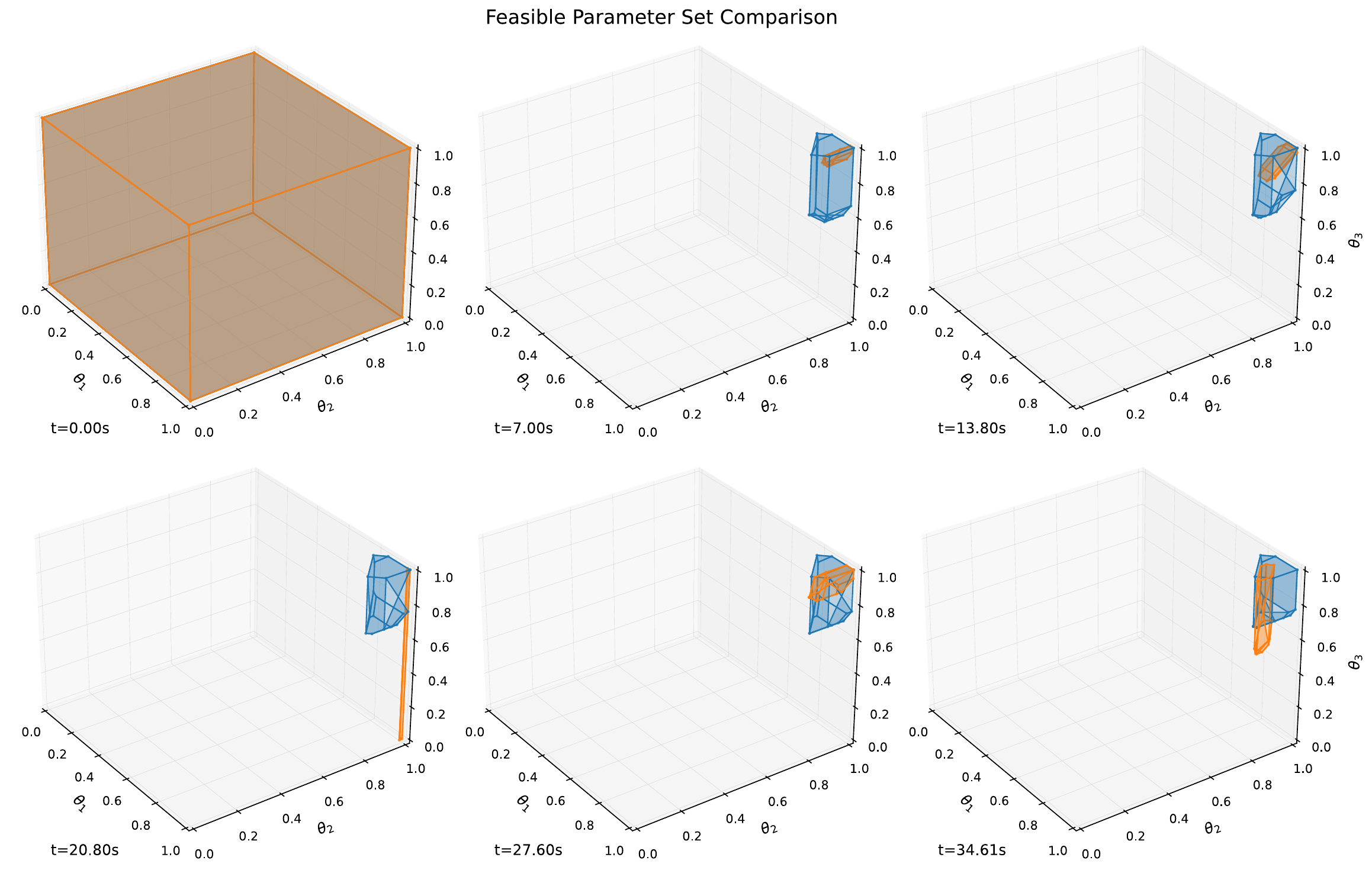}
  \caption{Evolution of the FPS in healthy conditions. In blue, the FPS considers measurement noise, converging towards the ``healthy" region. In contrast, the orange FPS, which neglects noise, becomes infeasible multiple times and fails to converge uniformly.}
  \label{fig:healthy_set}
\end{figure}

\begin{figure}[tbh!]
  \centering
  \includegraphics[width=\linewidth]{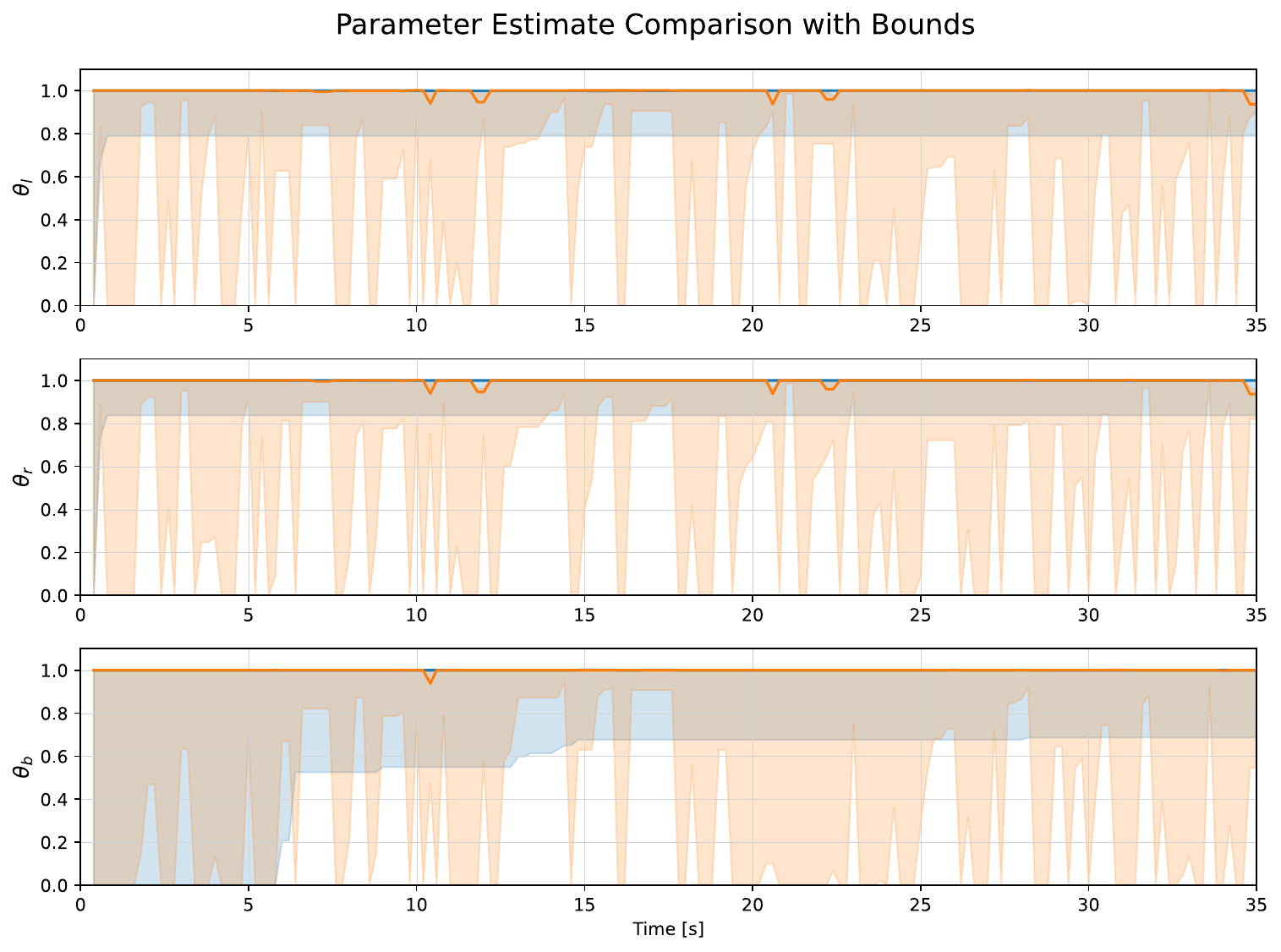}
  \caption{Evolution of the parameter estimate along with the corresponding bounds of the FPS projected in one dimension. The orange set, which does not account for measurement noise, becomes infeasible in several instances, triggering false alarms.}
  \label{fig:healthy_param}
\end{figure}

We begin by simulating the ASV following a sinusoidal reference path under healthy conditions to compare the formulation developed in this work with existing approaches that ignore noise bounds, leading to false alarms. Figure \ref{fig:healthy_set} illustrates the evolution of two different FPSs at six equidistant time instances. In blue, we represent the FPS using the proposed UPS formulation introduced here, which accounts for measurement noise. In orange, we show the FPS from the existing UPS formulation that does not consider noise. Our formulation ensures the FPS remains feasible in healthy conditions, converging towards a ``healthy area" around the healthy value $\bm{\theta} = \left( 1 \; 1 \; 1 \right)^{\top}$. In contrast, the previous formulation leads to several instances where the FPS becomes infeasible, preventing consistent convergence to the ``healthy area". This effect is even more pronounced in Figure \ref{fig:healthy_param}, where the time evolution of each fault parameter $\theta_i$ is shown, along with the corresponding set bounds as shaded areas of the same color. It is clear that the orange FPS in Figure \ref{fig:healthy_param} becomes infeasible multiple times, triggering false alarms. In contrast, our formulation is designed to prevent false alarms entirely and the set in blue converges monotonically to the healthy region.

\begin{figure}[tbh!]
  \centering
  \includegraphics[width=\linewidth]{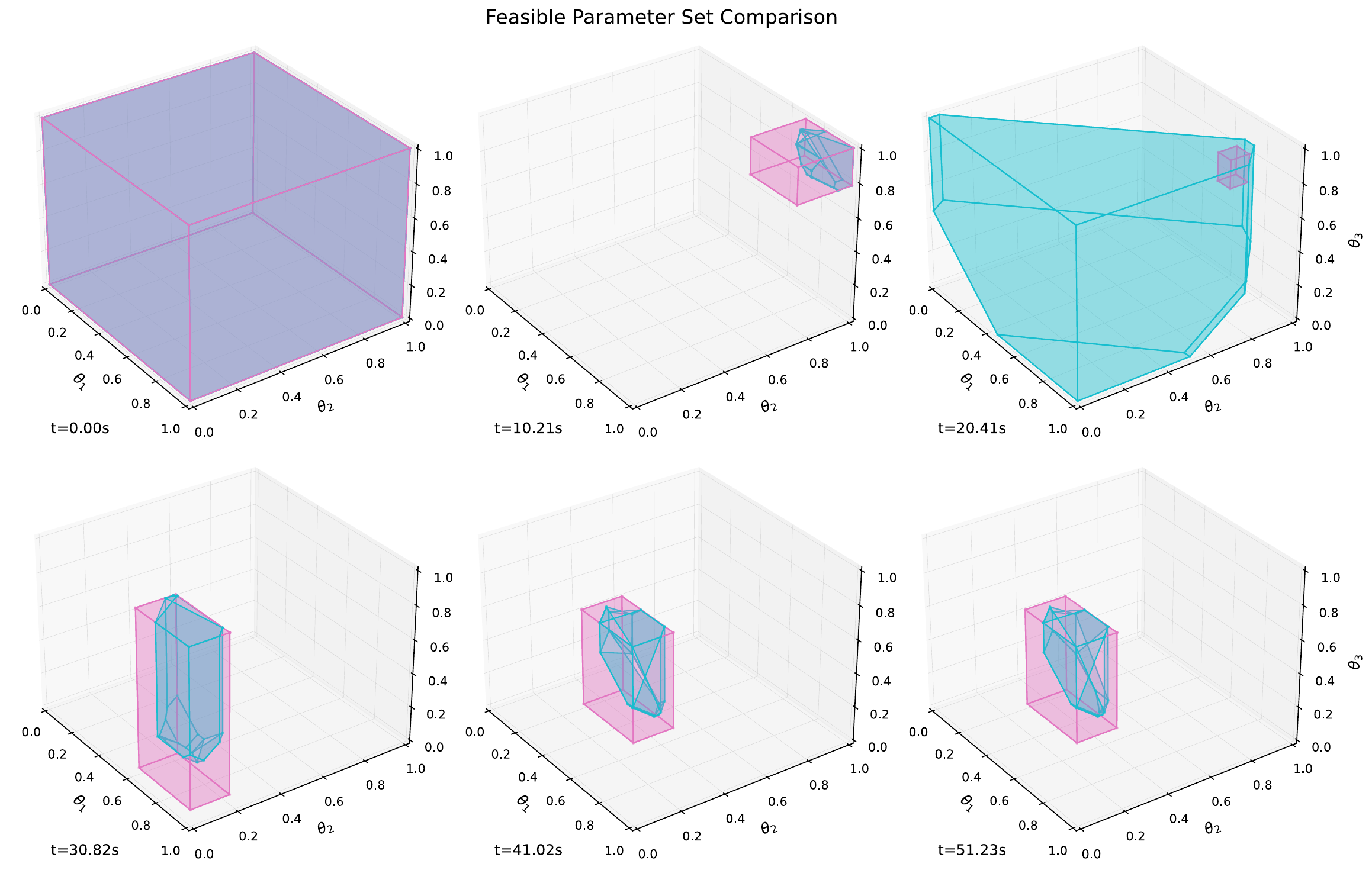}
  \caption{Comparison of the FPS evolution using two different outer approximations. The top-right sub-figure highlights the moment just after the fault occurs. The tighter outer approximation (cyan) detects the fault faster and begins to converge toward the faulty region, while the looser approximation (pink) converges more slowly.}
  \label{fig:faulty_set}
\end{figure}

\begin{figure}[tbh!]
  \centering
  \includegraphics[width=\linewidth]{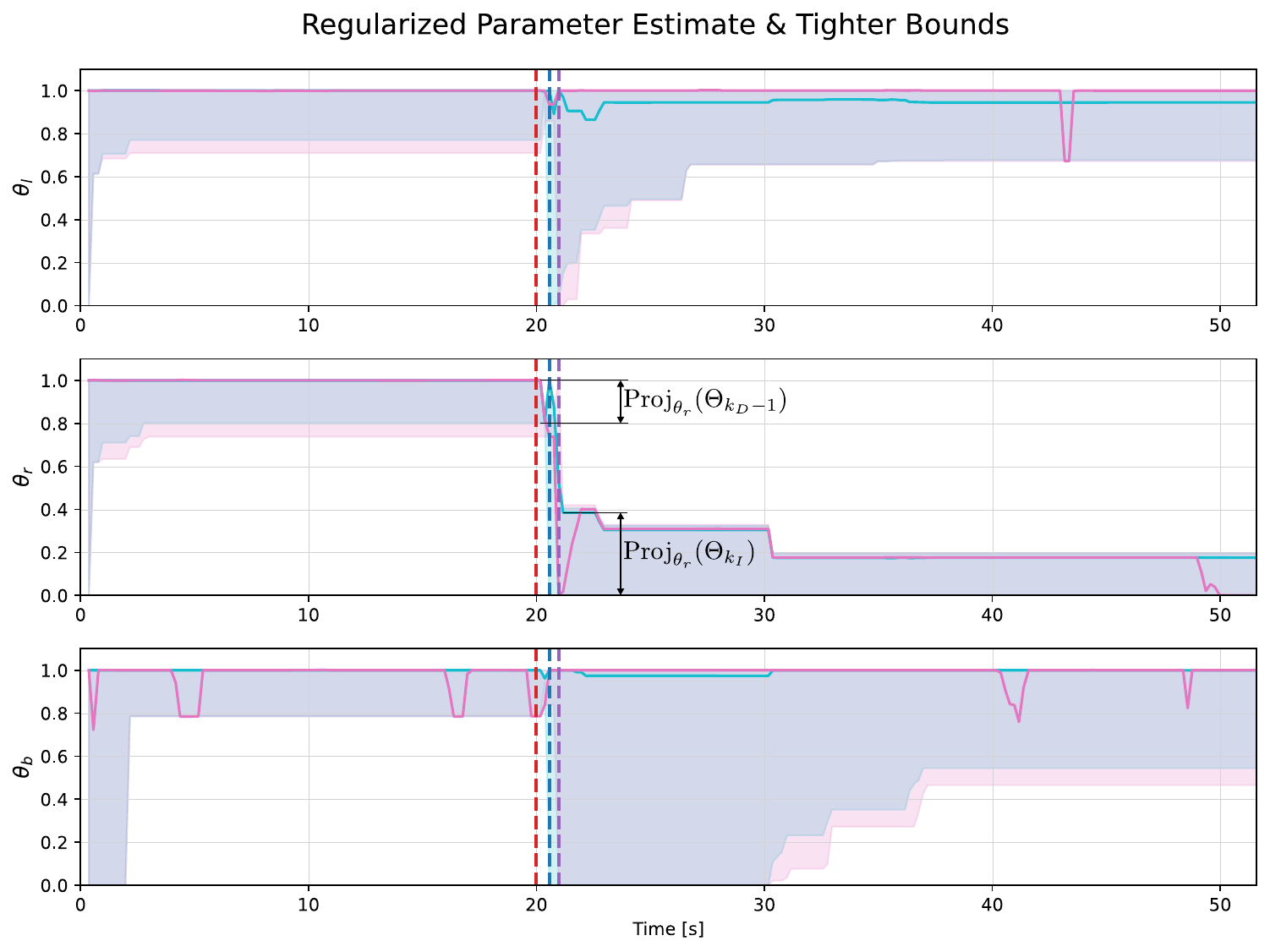}
  \caption{Parameter estimate with corresponding bounds (shaded areas). In the second sub-plot for $\theta_r$, the discontinuity in FPS bounds indicates the fault in the right thruster. The fault occurs at the red dashed vertical line, with the detection time shown in purple. The regularized estimate (cyan) is more stable and closer to the true value compared to the unregularized estimate (pink).}
  \label{fig:faulty_param}
\end{figure}

We also simulate a traffic scenario where the ASV needs to follow a straight, horizontal reference path while avoiding collisions with Obstacle Vessels (OVs). The system is subjected to both disturbance and noise and at time $t_F = \Delta t k_f= 20s$, a permanent fault $\theta_r = 0.2$ is injected in the right thruster. We compare different outer approximations of the FPS and evaluate their sensitivity, as well as compare the regularized parameter estimate proposed here with the conventional one. Both FPSs are constructed from UPSs that account for measurement noise. Figure \ref{fig:faulty_set} shows the evolution of the FPS using two different outer approximations. The cyan line represents a ``tighter" outer approximation from Algorithm \ref{alg:outer_approximation} with $\phi=1$, while the pink line illustrates a ``looser" approximation with $\phi=0$ (a simple bounding box). At time $t=20.41s$, after the fault occurs, it is evident that the tighter approximation in cyan, being more sensitive, leads to an infeasible set, indicating a fault as it starts converging towards a ``faulty area". The looser approximation in pink also converges towards the faulty region but with some delay due to its larger volume. Figure \ref{fig:faulty_param} shows the parameter estimates for this scenario. In cyan, the regularized parameter estimate is displayed, while the un-regularized estimate ($\bm{\Lambda} = \bm{0}$) is shown in pink. The corresponding FPS bounds are shown in matching colors. The moment of the fault is marked by a red dashed vertical line. The fault is detected at $t=20.56s$ using the tighter approximation (blue dashed vertical line) and at $t=21s$ using the looser approximation (purple dashed vertical line). The regularized parameter estimate (cyan continuous line) is generally more stable than the unregularized one (pink continuous line), making it a more reliable nominal estimate to evaluate the true fault value.

The evolution of the FPS along with the ASV trajectories for both simulation experiments can be viewed in animated form in the video available at \cite{tsolakis2024c}.


\section{Conclusions}\label{s:conclusions}
In this work, we introduced a Fault Diagnosis method using Set Membership Estimation for uncertain nonlinear systems that are linear in the fault parameters and subject to both state and output uncertainties. Our approach enhances fault diagnosis by addressing both uncertainty types, improving robustness and accuracy. It employs an inverse test for reliable fault detection and isolation, continuously refining a feasible set for fault parameter estimation. Adaptive regularization in the estimation process provides greater precision, especially when input-output data are sparse, supporting fault identifiability. For future work, we aim to leverage these results to improve our trajectory optimization framework, enhancing its robustness against faults and allowing it to operate more reliably in environments shared with human-operated vehicles.

%%%%%%%%%%%%%%%%%%%%%%%%%%%%%%%%%%%%%%%%%%%%%%%%%%%%%%%%%%%%%%%%%%%%%%%%%%%%%%%%


%%%%%%%%%%%%%%%%%%%%%%%%%%%%%%%%%%%%%%%%%%%%%%%%%%%%%%%%%%%%%%%%%%%%%%%%%%%%%%%%

\bibliographystyle{IEEEtran}
\bibliography{IEEEabrv,mybib}

\end{document}